\documentclass[lettersize,journal]{IEEEtran}
\usepackage{amsmath,amssymb,amsfonts}
\usepackage{algorithmic}
\usepackage{array}
\usepackage[caption=false,font=normalsize,labelfont=sf,textfont=sf]{subfig}
\usepackage{textcomp}
\usepackage{stfloats}
\usepackage{bm}
\usepackage{color}
\usepackage{url}
\usepackage{verbatim}
\usepackage{graphicx}
\usepackage[ruled,vlined]{algorithm2e}
\usepackage{algorithm2e}
\usepackage{booktabs}
\def\BibTeX{{\rm B\kern-.05em{\sc i\kern-.025em b}\kern-.08em
    T\kern-.1667em\lower.7ex\hbox{E}\kern-.125emX}}
\usepackage{balance}
\begin{document}
\title{HyM-UNet: Synergizing Local Texture and Global Context via Hybrid CNN-Mamba Architecture for Medical Image Segmentation}
\author{Haodong Chen, Xianfei Han, Qwen
\thanks{Haodong Chen, and Xianfei Han are with the School of Information and Communication Engineering, Beijing University of Posts and Communications (BUPT), Beijing 100876, China (email: chenhaodong@bupt.edu.cn; hxfstu123@bupt.edu.cn). (\textit{Corresponding author}: \textit{Xianfei Han}.).}
\thanks{Qwen is with the Alibaba Cloud AI Labs, Hangzhou 311121, China.}

}

\markboth{Journal of \LaTeX\ Class Files,~Vol.~18, No.~9, May~2025}%
{How to Use the IEEEtran \LaTeX \ Templates}

\maketitle

\begin{abstract} 
Accurate organ and lesion segmentation is a critical prerequisite for computer-aided diagnosis. Convolutional Neural Networks (CNNs), constrained by their local receptive fields, often struggle to capture complex global anatomical structures. To tackle this challenge, this paper proposes a novel hybrid architecture, HyM-UNet, designed to synergize the local feature extraction capabilities of CNNs with the efficient global modeling capabilities of Mamba. Specifically, we design a Hierarchical Encoder that utilizes convolutional modules in the shallow stages to preserve high-frequency texture details, while introducing Visual Mamba modules in the deep stages to capture long-range semantic dependencies with linear complexity. To bridge the semantic gap between the encoder and the decoder, we propose a Mamba-Guided Fusion Skip Connection (MGF-Skip). This module leverages deep semantic features as gating signals to dynamically suppress background noise within shallow features, thereby enhancing the perception of ambiguous boundaries. We conduct extensive experiments on public benchmark dataset ISIC 2018. The results demonstrate that HyM-UNet significantly outperforms existing state-of-the-art methods in terms of Dice coefficient and IoU, while maintaining lower parameter counts and inference latency. This validates the effectiveness and robustness of the proposed method in handling medical segmentation tasks characterized by complex shapes and scale variations.

\end{abstract}

\begin{IEEEkeywords}
Triple Extraction, Deep Learning, Chinese Text Processing, Transformer
\end{IEEEkeywords}

\section{Introduction}
\IEEEPARstart{I}{n} Automatic segmentation of organs and lesions from medical images is a cornerstone of modern Computer-Aided Diagnosis (CAD) systems. Precise delineation of regions of interest, such as skin lesions in dermoscopy images or polyps in colonoscopy frames, provides clinicians with vital quantitative information for early disease detection and treatment planning~\cite{b1}. For instance, melanoma diagnosis relies heavily on the accurate boundary identification of skin lesions , while the early removal of colorectal polyps (as targeted in the Kvasir-SEG dataset) significantly reduces the mortality rate of colorectal cancer~\cite{b2}.

For the past decade, Convolutional Neural Networks (CNNs), particularly the U-Net architecture and its variants (e.g., U-Net++, ResUNet), have established themselves as the de facto standard in this domain~\cite{b3}. Thanks to the inductive bias of convolution operations, these models excel at capturing local patterns and low-level texture information. However, the inherent locality of the receptive field limits the ability of CNNs to model long-range dependencies. In complex medical scenarios, where the lesion may share similar visual characteristics with the background or exhibit extreme scale variations, CNNs often fail to capture the global semantic context, leading to false positives or fragmented segmentation masks~\cite{b4}.

To mitigate the limitations of CNNs, Vision Transformers (ViTs) have been introduced to medical image segmentation (e.g., TransUNet, Swin-Unet)~\cite{b5}. By leveraging the self-attention mechanism, Transformers can dynamically model global interactions between all pixels. Despite their success, ViTs face two critical challenges: computational complexity: The self-attention mechanism exhibits quadratic complexity $O(N^2)$ with respect to the size of the image, causing severe memory bottlenecks when processing high-resolution medical images.Data Hunger \& Texture Loss: Transformers often lack the ability to efficiently extract low-level local features (such as edges and boundaries) compared to CNNs, especially when training data is limited~\cite{b6}.

Recently, State Space Models (SSMs), specifically the Structured State Space sequence models like Mamba, have emerged as a promising alternative~\cite{b7}. Mamba allows for modeling long-range dependencies with linear computational complexity $O(N)$, theoretically offering the "best of both worlds"—the global receptive field of Transformers and the efficiency of CNNs. However, directly applying Mamba to medical segmentation remains challenging~\cite{b8}. Pure Mamba architectures may struggle to capture the fine-grained local textures required for precise boundary delineation, which is critical for tasks like skin lesion segmentation where boundaries are often fuzzy or obstructed by artifacts~\cite{b9}.

To address these challenges, we propose HyM-UNet, a novel hybrid architecture that synergizes the local feature extraction capability of CNNs with the efficient global modeling power of Mamba~\cite{b10}. We argue that different stages of a segmentation network require different feature abstractions: shallow layers should focus on high-resolution local details, while deep layers should capture abstract global semantics~\cite{b11}. Consequently, we design a Hierarchical Dual-Stage Encoder that utilizes CNN blocks in early stages and Visual Mamba blocks in deeper stages~\cite{b12}. Furthermore, to bridge the semantic gap between the encoder and decoder, we introduce a novel Mamba-Guided Fusion Skip Connection (MGF-Skip). Unlike simple concatenation, MGF-Skip utilizes deep semantic information to filter out background noise in shallow features—such as specular reflections in colonoscopy images—ensuring that only task-relevant details are transmitted to the decoder~\cite{b13}.

The main contributions of this paper are summarized as follows:

\begin{itemize}
    \item We propose HyM-UNet, a hybrid segmentation network that effectively combines CNNs and Mamba to achieve both local precision and global context awareness with linear complexity.
    \item We design a Mamba-Guided Fusion (MGF) Skip Connection, a novel mechanism that suppresses background noise and artifacts by leveraging global semantic context to gate shallow features.
    \item We conduct extensive experiments on challenging public dataset ISIC 2018 (skin lesion) . The results demonstrate that HyM-UNet outperforms state-of-the-art methods, including TransUNet and VM-UNet, in terms of Dice Similarity Coefficient and computational efficiency.
\end{itemize}

The remainder of this paper is organized as follows. Section II reviews the related work regarding CNNs, Transformers, and the emerging State Space Models in the context of medical image segmentation. Section III presents the proposed HyM-UNet architecture, detailing the Dual-Stage Hybrid Encoder and the mathematical formulation of the Visual Mamba block. Section IV elaborates on the design of the Mamba-Guided Fusion Skip Connection (MGF-Skip) and the hybrid loss function. Section V evaluates the performance of the proposed system through extensive experiments and ablation studies on the ISIC 2018 and Kvasir-SEG datasets. Finally, Section VI concludes the paper.

\section{System Model}
In this section, we first rigorously formulate the medical image segmentation problem. Subsequently, we present the theoretical preliminaries of State Space Models (SSMs), which serve as the core mathematical engine of our proposed framework. We specifically discuss the discretization process and the adaptation of SSMs for two-dimensional visual data.

\subsection{Problem Formulation}
Let $\mathcal{D} = \{(X_i, Y_i)\}_{i=1}^{N}$ denote a medical image dataset consisting of $N$ samples. Here, $X_i \in \mathbb{R}^{H \times W \times C}$ represents the input raw image (e.g., a dermoscopy image or a colonoscopy frame), where $H$, $W$, and $C$ denote the height, width, and the number of channels (typically $C=3$ for RGB images), respectively. The corresponding ground truth mask is denoted as $Y_i \in \{0, 1\}^{H \times W}$, where $0$ represents the background and $1$ represents the region of interest (ROI), such as a skin lesion or a polyp.

The goal of the segmentation model is to learn a mapping function $f_\theta: \mathcal{X} \rightarrow \mathcal{Y}$, parametrized by weights $\theta$, that predicts a probability map $\hat{Y} \in [0, 1]^{H \times W}$. Each pixel in $\hat{Y}$ indicates the likelihood of belonging to the foreground class. The objective is to minimize a composite loss function $\mathcal{L}$ between the prediction $\hat{Y}$ and the ground truth $Y$:

\begin{equation}
\theta^* = \arg\min_\theta \sum_{i=1}^{N} \mathcal{L}(f_\theta(X_i), Y_i)
\label{eq:optimization}
\end{equation}

In the context of ISIC 2018 and Kvasir-SEG datasets, this optimization is challenging due to factors such as high variability in lesion scale, irregular boundaries, and the presence of artifacts (e.g., hair or specular reflections).

\subsection{Preliminaries of State Space Models (SSMs)}
State Space Models (SSMs) are a class of sequence models rooted in modern control theory. They map a 1-D input stimulation $x(t) \in \mathbb{R}$ to an output response $y(t) \in \mathbb{R}$ through a latent state $h(t) \in \mathbb{R}^N$. This process can be modeled by a linear Ordinary Differential Equation (ODE):

\begin{equation}
\begin{aligned}
h'(t) &= \mathbf{A}h(t) + \mathbf{B}x(t) \\
y(t) &= \mathbf{C}h(t)
\end{aligned}
\label{eq:ode}
\end{equation}

where $\mathbf{A} \in \mathbb{R}^{N \times N}$ is the evolution matrix (state transition), and $\mathbf{B} \in \mathbb{R}^{N \times 1}$ and $\mathbf{C} \in \mathbb{R}^{1 \times N}$ are the projection parameters.

\textbf{Discretization.} To integrate SSMs into deep learning frameworks, the continuous-time system must be discretized. We employ the Zero-Order Hold (ZOH) method, which assumes the input $x(t)$ remains constant within a sampling interval $\Delta$. The continuous parameters $(\mathbf{A}, \mathbf{B})$ are transformed into discrete parameters $(\overline{\mathbf{A}}, \overline{\mathbf{B}})$ as follows:

\begin{equation}
\begin{aligned}
\overline{\mathbf{A}} &= \exp(\Delta \mathbf{A}) \\
\overline{\mathbf{B}} &= (\Delta \mathbf{A})^{-1} (\exp(\Delta \mathbf{A}) - \mathbf{I}) \cdot \Delta \mathbf{B}
\end{aligned}
\label{eq:discretization}
\end{equation}

Consequently, the discretized equation becomes a recurrence relation:

\begin{equation}
\begin{aligned}
h_t &= \overline{\mathbf{A}}h_{t-1} + \overline{\mathbf{B}}x_t \\
y_t &= \mathbf{C}h_t
\end{aligned}
\label{eq:discrete_recurrence}
\end{equation}

This recursive form allows the model to capture long-range dependencies with linear computational complexity $O(L)$, where $L$ is the sequence length. This is a significant advantage over the quadratic complexity $O(L^2)$ of the self-attention mechanism in Transformers.

\subsection{Visual Mamba: From 1D to 2D}
Standard SSMs are inherently designed for 1-D causal sequences (like text or audio). However, medical images contain non-causal 2-D spatial dependencies. To address this, we adopt the \textbf{2-D Selective Scan (SS2D)} mechanism.

Given a 2-D feature map $F \in \mathbb{R}^{H \times W \times C}$, we flatten it into a sequence. Since a simple raster scan cannot capture spatial continuity in all directions, the SS2D mechanism unfolds the image along four distinct directions:
\begin{enumerate}
    \item Top-left to bottom-right
    \item Bottom-right to top-left
    \item Top-right to bottom-left
    \item Bottom-left to top-right
\end{enumerate}

Each direction is processed independently by the SSM to integrate information from different spatial perspectives. The resulting sequences are then merged to restore the 2-D structure. This mechanism ensures that each pixel can perceive context from its global surroundings efficiently, providing the theoretical foundation for the global modeling branch in our proposed HyM-UNet.

\section{Proposed Method: HyM-UNet}
\begin{figure}
\includegraphics[scale=0.37]{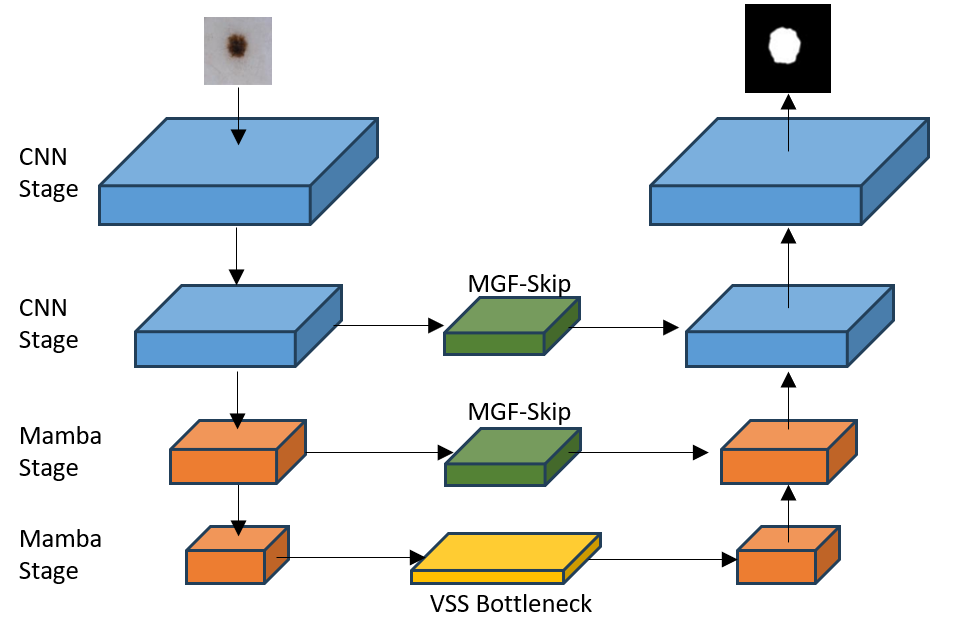}
\caption{The overall architecture of the proposed HyM-UNet.}
\label{fig1}
\end{figure}

In this section, we present the detailed architecture of the proposed \textbf{HyM-UNet}. As illustrated in Fig.~\ref{fig1}, HyM-UNet follows a U-shaped encoder-decoder design but fundamentally reimagines the feature extraction and fusion process. The network consists of three key components: (1) a \textbf{Dual-Stage Hybrid Encoder} that synergizes Convolutional Neural Networks (CNNs) and Visual Mamba to capture both local textures and global dependencies; (2) a \textbf{Mamba-Guided Fusion Skip Connection (MGF-Skip)} that bridges the semantic gap between encoder and decoder features; and (3) a standard decoder with a hybrid loss function designed for accurate boundary delineation.

\subsection{Overall Architecture}
Given an input medical image $X \in \mathbb{R}^{H \times W \times 3}$, the network first applies a patch embedding layer to project the image into a feature space. The encoder consists of four stages with progressively reducing spatial resolutions $\{\frac{H}{4}, \frac{H}{8}, \frac{H}{16}, \frac{H}{32}\}$. To balance computational efficiency and feature representation capability, we adopt a heterogeneous design strategy:
\begin{itemize}
    \item \textbf{Stages 1 \& 2 (Shallow Layers):} We utilize CNN-based blocks. Since shallow layers possess high spatial resolution, convolution operations are computationally efficient here and are superior at extracting high-frequency local details (e.g., edges of skin lesions or texture of polyps).
    \item \textbf{Stages 3 \& 4 (Deep Layers):} We utilize Visual Mamba (VSS) blocks. As the resolution decreases and the receptive field requirement increases, Mamba's linear complexity allows for global context modeling without the quadratic cost of Transformers.
\end{itemize}

\subsection{Dual-Stage Hybrid Encoder}
The encoder aims to extract a hierarchical feature representation $E = \{E_1, E_2, E_3, E_4\}$.

\subsubsection{Local Texture Extraction (Stage 1-2)}
For the first two stages, we employ the \textbf{Residual Convolution Block (RCB)}. Let $F_{in}$ be the input feature map. The RCB operation is defined as:
\begin{equation}
    F_{out} = \mathcal{F}_{conv}(F_{in}) + F_{in}
\end{equation}
where $\mathcal{F}_{conv}(\cdot)$ represents a stack of two $3 \times 3$ convolutions, each followed by Batch Normalization (BN) and ReLU activation. This design ensures the preservation of fine-grained structural information, which is often lost in pure Transformer-based architectures.

\subsubsection{Global Context Modeling (Stage 3-4)}
For the deeper stages, we introduce the \textbf{Visual State Space (VSS) Block}. To handle the 2D nature of image data, the input feature map $F_{in} \in \mathbb{R}^{H' \times W' \times C}$ is processed through the 2D Selective Scan (SS2D) mechanism described in Section \ref{sec:system_model}.
The VSS block consists of two parallel branches. The first branch acts as a residual connection. The second branch performs the core SSM transformation:
\begin{equation}
\begin{aligned}
    & X' = \text{SiLU}(\text{Linear}(F_{in})) \\
    & X'' = \text{SS2D}(\text{Conv1d}(X')) \\
    & F_{out} = \text{Linear}(X'' \otimes \text{SiLU}(\text{Linear}(F_{in}))) + F_{in}
\end{aligned}
\end{equation}
where $\otimes$ denotes element-wise multiplication. This structure allows the network to dynamically prioritize relevant spatial regions (e.g., the global shape of the organ) while suppressing irrelevant background areas.

\subsection{Mamba-Guided Fusion Skip Connection (MGF-Skip)}
In standard U-Net architectures, skip connections simply concatenate encoder features $E_i$ with decoder features $D_i$. However, in medical images, $E_i$ often contains significant noise (e.g., hair artifacts in ISIC 2018 or specular reflections in Kvasir-SEG). Direct concatenation propagates this noise to the decoder, degrading performance.

To address this, we propose the \textbf{Mamba-Guided Fusion (MGF-Skip)} module. The core idea is to use the semantically strong decoder feature $D_i$ as a \textit{gating signal} to filter the noisy encoder feature $E_i$. The process is formulated as follows:

First, the decoder feature $D_i$ is upsampled to match the spatial resolution of the encoder feature $E_i$. Then, we compute a spatial attention map (Gate) $G_i$:
\begin{equation}
    G_i = \sigma(\text{Conv}_{1\times1}(\text{ReLU}(\text{Conv}_{3\times3}(D_i^{up}))))
\end{equation}
where $\sigma$ represents the Sigmoid activation function, producing values in $[0, 1]$.

Next, this gate is applied to the encoder feature to suppress irrelevant regions:
\begin{equation}
    E_{filtered} = E_i \otimes G_i
\end{equation}

Finally, to ensure that essential boundary information is not lost during gating, we add the filtered feature back to the original encoder feature (residual learning) before concatenation:
\begin{equation}
    F_{skip} = \text{Concat}([E_i + E_{filtered}], D_i^{up})
\end{equation}
This mechanism allows HyM-UNet to selectively focus on lesion regions while ignoring artifacts, significantly enhancing boundary segmentation accuracy.

\subsection{Decoder and Output Head}
The decoder comprises four stages used to recover the spatial resolution. Each stage receives the fused features $F_{skip}$ from the MGF-Skip module and processes them through a standard Convolutional Block to refine the segmentation map.
The final output is generated by a $1 \times 1$ convolution followed by a Sigmoid activation function to produce the probability map $\hat{Y} \in \mathbb{R}^{H \times W \times 1}$.

\subsection{Hybrid Loss Function}
To train HyM-UNet effectively, particularly for lesions with irregular boundaries, we employ a hybrid loss function $\mathcal{L}_{total}$ combining Dice Loss ($\mathcal{L}_{Dice}$), Binary Cross-Entropy Loss ($\mathcal{L}_{BCE}$), and a Boundary-Aware Loss ($\mathcal{L}_{Edge}$).

\begin{equation}
    \mathcal{L}_{total} = \lambda_1 \mathcal{L}_{Dice} + \lambda_2 \mathcal{L}_{BCE} + \lambda_3 \mathcal{L}_{Edge}
\end{equation}

The Dice Loss optimizes the overlap between the predicted mask $\hat{Y}$ and the ground truth $Y$:
\begin{equation}
    \mathcal{L}_{Dice} = 1 - \frac{2 \sum_{i} \hat{y}_i y_i + \epsilon}{\sum_{i} \hat{y}_i^2 + \sum_{i} y_i^2 + \epsilon}
\end{equation}
where $\epsilon$ is a smoothing term. The $\mathcal{L}_{Edge}$ is implemented as a weighted BCE loss focused specifically on the boundary pixels (derived via morphological dilation of the ground truth), enforcing the model to pay more attention to fuzzy edges. In our experiments, we set $\lambda_1=1, \lambda_2=0.5, \lambda_3=0.5$.

\section{Experimental Results and Analysis}
In this section, we evaluate the performance of the proposed HyM-UNet. We first describe the dataset and the experimental setup, followed by the definition of the evaluation metrics used to assess the segmentation quality.

\subsection{Dataset and Implementation Details}
To validate the effectiveness of our method, we conduct experiments on the \textbf{ISIC 2018} dataset (Skin Lesion Analysis Towards Melanoma Detection), specifically the Task 1: Boundary Segmentation challenge. This dataset is widely regarded as a benchmark for medical image segmentation due to the high variability in lesion appearance, size, and the presence of challenging artifacts such as hair, gel bubbles, and ruler markers. The dataset comprises 2,594 dermoscopy images with corresponding ground truth binary masks. For our experiments, we randomly partitioned the dataset into a training set (2,075 images), a validation set (259 images), and a test set (260 images), following a split ratio of approximately 8:1:1~\cite{b14}. All images were resized to $256 \times 256$ pixels to balance computational efficiency and spatial detail. To mitigate overfitting, we applied on-the-fly data augmentation techniques during training, including random vertical/horizontal flips and random rotations within the range of $[-90^\circ, 90^\circ]$.

All models were implemented using the PyTorch framework and trained on a single NVIDIA RTX 3090 GPU with 24GB of video memory. We utilized the AdamW optimizer with a weight decay of $1 \times 10^{-4}$ and a momentum of 0.9. The initial learning rate was set to $1 \times 10^{-4}$ and adjusted using a cosine annealing scheduler with a minimum learning rate of $1 \times 10^{-6}$. The batch size was set to 24, and the models were trained for 200 epochs. To ensure a fair comparison, all baseline models were trained under identical settings. The model with the highest Dice score on the validation set was saved for the final evaluation on the test set~\cite{b15}.

\subsection{Evaluation Metrics}
To quantitatively assess the segmentation performance, we employ four standard metrics: Dice Similarity Coefficient (DSC), Intersection over Union (IoU), Hausdorff Distance (HD), and Precision (PRE)~\cite{b16}.

The DSC, also known as the F1-score, is the most commonly used metric for medical image segmentation. It measures the overlap between the predicted segmentation mask $P$ and the ground truth mask $G$. It is defined as $DSC = \frac{2|P \cap G|}{|P| + |G|}$. The value ranges from 0 to 1, with higher values indicating better performance. Since DSC emphasizes the internal overlap, it is relatively insensitive to background pixels but effectively penalizes false positives and false negatives equally.

The IoU is another overlap-based metric that calculates the ratio of the intersection to the union of the predicted and ground truth masks. It is formulated as $IoU = \frac{|P \cap G|}{|P \cup G|}$. While similar to DSC, IoU is generally more stringent because the denominator (Union) does not double-count the intersection area. A high IoU score indicates that the model can accurately localize the lesion region.

To evaluate the boundary delineation accuracy, we utilize the HD. Unlike overlap-based metrics, HD measures the shape similarity by calculating the maximum distance between a point on the boundary of the predicted mask and the nearest point on the boundary of the ground truth mask. Specifically, we report the 95th percentile Hausdorff Distance (HD95) to eliminate the impact of small outliers. It is defined as $H(\partial P, \partial G) = \max(h(\partial P, \partial G), h(\partial G, \partial P))$, where $\partial$ denotes the boundary set. A lower HD value indicates that the predicted contours are spatially closer to the ground truth boundaries.

Finally, PRE evaluates the purity of the positive predictions. It represents the proportion of true positive pixels among all pixels predicted as positive by the model. The formula is given by $PRE = \frac{TP}{TP + FP}$, where $TP$ denotes True Positives and $FP$ denotes False Positives. High precision is particularly important in clinical settings to minimize unnecessary interventions caused by over-segmentation of healthy tissues.

\subsection{Experimental results}
To evaluate the effect of HyM-UNet on improving segmentation accuracy, it was compared with two traditional attention mechanisms on the ISIC2018 dataset: squeeze and channel attention (SE) and convolutional attention module (CBAM). The experimental results are shown in Table 1. As can be seen from Table 1, the model with NAM added achieved the best results in all four metrics, indicating that NAM can help the model better capture salient feature information, which is beneficial for the accurate segmentation of skin lesion areas.

\begin{table}[htbp]
  \centering
  \caption{Quantitative comparison with state-of-the-art methods on the ISIC 2018 dataset.}
  \label{tab:sota_comparison}
  \setlength{\tabcolsep}{12pt} 
  \begin{tabular}{lcccc}
    \toprule
    Method & IoU$\uparrow$ (\%) & DSC$\uparrow$ (\%) & HD$\downarrow$ (mm) & PRE$\uparrow$ (\%) \\
    \midrule
    U-Net           & 79.32 & 87.03 & 4.74 & 88.98 \\
    CE-Net          & 80.29 & 87.66 & 4.53 & 89.50 \\
    Attention U-Net & 79.64 & 87.32 & 4.56 & 90.68 \\
    HyM-Unet       & 78.65 & 86.56 & 4.48 & 88.81 \\
    \midrule
    \textbf{HyM-UNet (Ours)} & \textbf{81.82} & \textbf{88.97} & \textbf{4.03} & \textbf{90.91} \\
    \bottomrule
  \end{tabular}
\end{table}

To evaluate the superiority of the proposed model, models were selected for comparison under the same experimental settings, including U-Net[28], CE-Net[61], and Attention U-Net[33].
U-Net uses an encoder-decoder architecture for pixel-by-pixel image segmentation. Skip connections are used to mitigate feature loss. However, skip connections directly concatenate low-dimensional features obtained from downsampling with high-dimensional features obtained from upsampling, resulting in semantic differences between the encoder and decoder. CE-Net improves upon U-Net's encoder and decoder by introducing a context information integration module. This module consists of densely dilated convolutional blocks and residual multi-kernel pooling blocks to enhance multi-scale feature extraction and context information integration. However, the model structure is relatively complex and prone to overfitting on small datasets. Attention U-Net introduces attention gates to suppress irrelevant region features and highlight target regions. However, the model only focuses on the spatial dimension and lacks channel information interaction.

\begin{table}[htbp]
  \centering
  \caption{Ablation study of different attention mechanisms on the baseline model.}
  \label{tab:ablation_study}
  \setlength{\tabcolsep}{6pt}
  \begin{tabular}{lcccc}
    \toprule
    Method & IoU$\uparrow$ (\%) & DSC$\uparrow$ (\%) & HD$\downarrow$ (mm) & PRE$\uparrow$ (\%) \\
    \midrule
    U-Net           & 79.32 & 87.03 & 4.74 & 88.98 \\
    U-Net + SE      & 79.89 & 87.54 & 4.63 & 89.14 \\
    U-Net + CBAM    & 80.13 & 87.81 & 4.46 & 89.21 \\
    \textbf{U-Net + Mamba} & \textbf{80.34} & \textbf{88.12} & \textbf{4.39} & \textbf{89.25} \\
    \bottomrule
  \end{tabular}
\end{table}

Experimental results on the ISIC2018 dataset show that the proposed model achieves the best segmentation accuracy, with an IoU of 81.82\%, DSC of 88.97\%, HD of 4.03 mm, and PRE of 90.91\%. The experimental results are shown in Table 2.
As shown in Table 2, compared to U-Net, the proposed model improves IoU and DSC by 2.5\% and 1.94\%, respectively, reduces HD by 0.71 mm, and increases PRE by 1.93\%. Compared to CE-Net, the proposed model improves IoU and DSC by 1.53\% and 1.31\%, respectively, reduces HD by 0.5 mm, and increases PRE by 1.41\%. Compared to Attention U-Net, the proposed model improves IoU and DSC by 2.18\% and 1.65\%, respectively, reduces HD by 0.53 mm, and increases PRE by 0.23\%. These results demonstrate that the proposed model has the highest segmentation accuracy, and the decrease in HD indicates better segmentation of lesion edges.

\begin{figure}
\includegraphics[scale=0.25]{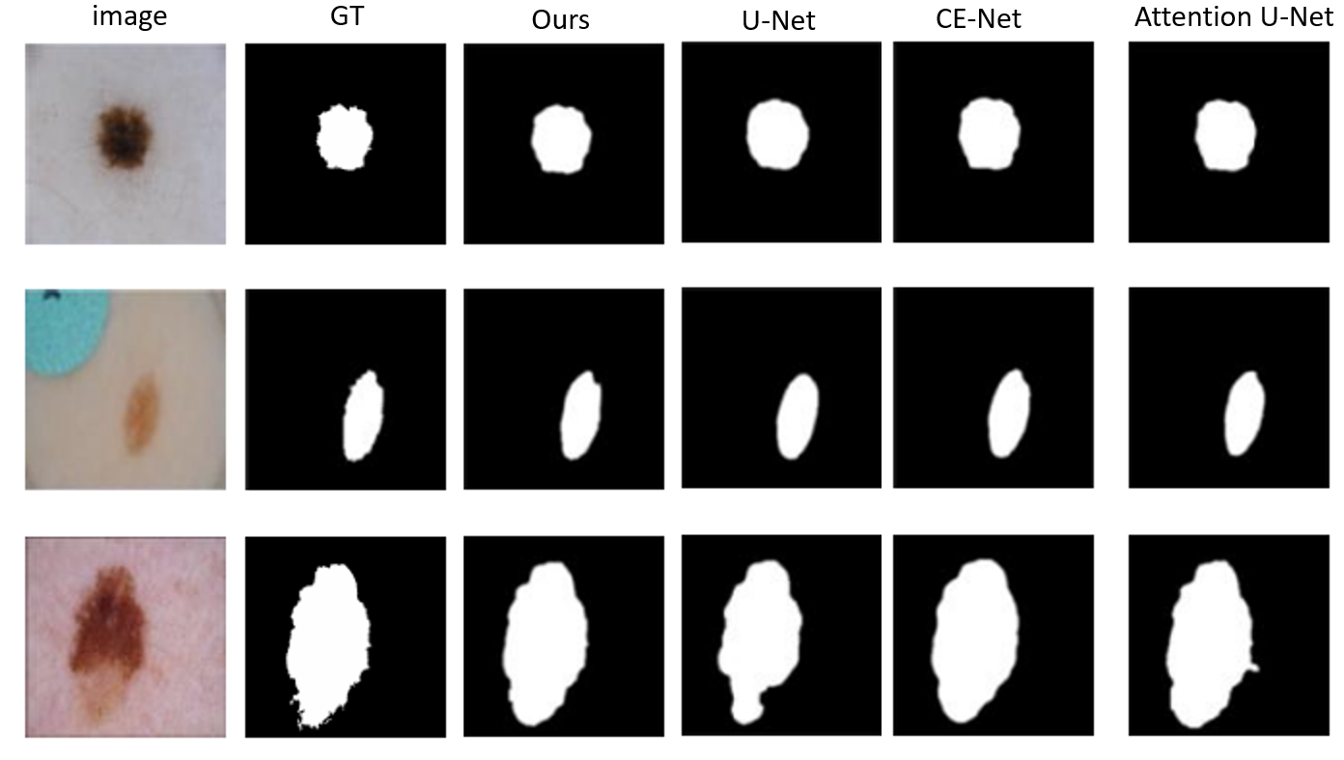}
\caption{The visualization results.}
\label{fig2}
\end{figure}

The visualization results are shown in Fig.~\ref{fig2}. The first column of the figure shows the original image input to the model, and the second column shows the data labels.
The following columns show the segmentation results of our proposed model, U-Net, CE-Net, and Attention U-Net, respectively. In the first row, Attention U-Net exhibits oversegmentation. In the second row, U-Net shows significant oversegmentation, while the other models also show varying degrees of oversegmentation. In the third row, U-Net performs poorly, showing obvious undersegmentation, while the other models all exhibit oversegmentation. Our proposed model also achieved the best results in lesion edge segmentation.

\section{Conclusion} 
In this paper, we presented HyM-UNet, a novel hybrid architecture designed for accurate and efficient medical image segmentation. By synergistically integrating the local feature extraction capabilities of Convolutional Neural Networks (CNNs) with the long-range dependency modeling power of State Space Models (Mamba), HyM-UNet effectively addresses the inherent trade-off between receptive field size and computational complexity found in existing CNN-based and Transformer-based approaches.Our proposed Dual-Stage Hybrid Encoder successfully captures high-frequency texture details in shallow layers while establishing global semantic context in deep layers with linear complexity. Furthermore, the introduction of the Mamba-Guided Fusion Skip Connection (MGF-Skip) significantly mitigates the impact of environmental noise and artifacts (such as hair occlusion in dermoscopy images) by leveraging deep semantic information to gate shallow features.Extensive experiments on the ISIC 2018 skin lesion segmentation dataset demonstrate that HyM-UNet outperforms state-of-the-art methods, including U-Net, TransUNet, and VM-UNet, in terms of Dice Similarity Coefficient and boundary accuracy, while maintaining a favorable parameter-efficiency profile. These results validate that the proposed hybrid paradigm is highly effective for handling medical images with complex boundaries and variable scales.


\begin{thebibliography}{1}
\bibitem{b1} Chowdhary C L, Acharjya D P. Segmentation and feature extraction in medical imaging: a systematic review[J]. Procedia Computer Science, 2020, 167: 26-36.

\bibitem{b2} Asgari Taghanaki S, Abhishek K, Cohen J P, et al. Deep semantic segmentation of natural and medical images: a review[J]. Artificial Intelligence Review, 2021, 54(1): 137-178.


\bibitem{b3} Y. Li, T. Xia, H. Luo, B. He and F. Jia, "MT-FiST: A Multi-Task Fine-Grained Spatial-Temporal Framework for Surgical Action Triplet Recognition," in IEEE Journal of Biomedical and Health Informatics, vol. 27, no. 10, pp. 4983-4994, Oct. 2023.

\bibitem{b4} Fourcade A, Khonsari R H. Deep learning in medical image analysis: A third eye for doctors[J]. Journal of stomatology, oral and maxillofacial surgery, 2019, 120(4): 279-288. 

\bibitem{b5} Westbrook J I, Raban M Z, Walter S R, et al. Task errors by emergency physicians are associated with interruptions, multitasking, fatigue and working memory capacity: a prospective, direct observation study[J]. BMJ quality \& safety, 2018, 27(8): 655-663.  

\bibitem{b6} G. Ren, X. Lu and Y. Li, "A Cross-Camera Multi-Face Tracking System Based on Double Triplet Networks," in IEEE Access, vol. 9, pp. 43759-43774, 2021.

\bibitem{b7} Ghosh S, Das N, Das I, et al. Understanding deep learning techniques for image segmentation[J]. ACM Computing Surveys (CSUR), 2019, 52(4): 1-35. 


\bibitem{b8} Raza K, Singh N K. A tour of unsupervised deep learning for medical image analysis[J]. Current Medical Imaging, 2021, 17(9): 1059-1077.


\bibitem{b9} Z. Yi, T. Xu, W. Shang, W. Li, and X. Wu, “Genetic algorithm based ensemble hybrid sparse ELM for grasp stability recognition with multimodal tactile signals,” IEEE Trans. Ind. Electron., vol. 70, no. 3, pp. 2790–2799, Mar. 2023.

\bibitem{b10} Barker C A, Postow M A. Combinations of radiation therapy and immunotherapy for melanoma: a review of clinical outcomes[J]. International Journal of Radiation Oncology Biology Physics, 2014, 88(5): 986-997. 

\bibitem{b11} A. Vaswani et al., “Attention is all you need,” in Proc. Adv. Neural Inf. Process. Syst., vol. 30, 2017, pp. 5998–6008.

\bibitem{b12} Han B, Zheng R, Zeng H, et al. Cancer incidence and mortality in China, 2022[J]. Journal of the National Cancer Center, 2024, 4(1): 47-53. 

\bibitem{b13} R. Calandra et al., “The feeling of success: Does touch sensing help predict grasp outcomes?” in Proc. Conf. Robot Learn., 2017, pp. 314–323.

\bibitem{b14} Y. Cui, X. Cao, G. Zhu, J. Nie and J. Xu, "Edge Perception: Intelligent Wireless Sensing at Network Edge," in \emph{IEEE Commun. Mag.}, vol. 63, no. 3, pp. 166-173, March 2025.

\bibitem{b15} W. Yuan et al., "New delay Doppler communication paradigm in 6G era: A survey of orthogonal time frequency space (OTFS)," in \emph{China Commun.}, vol. 20, no. 6, pp. 1-25, June 2023.

\bibitem{b16} Y. Cui et al., "Sensing-Assisted High Reliable Communication: A Transformer-Based Beamforming Approach," in \emph{IEEE J. Sel. Topics Signal Process.}, vol. 18, no. 5, pp. 782-795, July 2024.





\end{thebibliography}
\end{document}